\documentclass[conference]{IEEEtran}
\IEEEoverridecommandlockouts

\usepackage{cite}
\usepackage{amsmath,amssymb,amsfonts}
\usepackage{graphicx}

\usepackage{textcomp}
\usepackage{xcolor}
\def\BibTeX{{\rm B\kern-.05em{\sc i\kern-.025em b}\kern-.08em
    T\kern-.1667em\lower.7ex\hbox{E}\kern-.125emX}}

\usepackage{booktabs}
\usepackage{algorithm,algpseudocode}
\usepackage{newfloat}
\usepackage{listings}
\usepackage{xcolor}
\usepackage{amsmath}
\usepackage{amssymb}
\usepackage{enumitem}
\newlist{checklist}{itemize}{1}
\setlist[checklist,1]{
  label={\(\square\)},
  labelwidth=1em,
  labelsep=1em,
  left=0pt,
  itemsep=0pt
}

\usepackage{cite}
\usepackage{amsmath,amssymb,amsfonts}
\usepackage{graphicx}
\usepackage{textcomp}
\usepackage{xcolor}
\def\BibTeX{{\rm B\kern-.05em{\sc i\kern-.025em b}\kern-.08em
    T\kern-.1667em\lower.7ex\hbox{E}\kern-.125emX}}

\IEEEoverridecommandlockouts

\author{\IEEEauthorblockN{ Jiechao Gao* \thanks{* Corresponding author: Jiechao Gao}}
\IEEEauthorblockA{\textit{Department of Computer Science} \\
\textit{University of Virginia}\\
Charlottesville, VA, USA\\
jg5ycn@virginia.edu}
\and
\IEEEauthorblockN{Yuangang Li}
\IEEEauthorblockA{\textit{Thomas Lord Department of Computer Science} \\
\textit{University of Southern California}\\
Los Angeles, US \\
yuangang@usc.edu}
\and
\IEEEauthorblockN{Syeda Faiza Ahmed}
\IEEEauthorblockA{\textit{Department of Computer Science} \\
\textit{University of Dhaka}\\
Bangladesh \\
s2017115067@rme.du.ac.bd}

}

\title{Fed-LDR: Federated Local Data-infused Graph Creation with Node-centric Model Refinement}

\begin{document}

\maketitle

\begin{abstract}
The rapid acceleration of global urbanization has introduced novel challenges in enhancing urban infrastructure and services. Spatio-temporal data, integrating spatial and temporal dimensions, has emerged as a critical tool for understanding urban phenomena and promoting sustainability. In this context, Federated Learning (FL) has gained prominence as a distributed learning paradigm aligned with the privacy requirements of urban IoT environments. However, integrating traditional and deep learning models into the FL framework poses significant challenges, particularly in capturing complex spatio-temporal dependencies and adapting to diverse urban conditions.
To address these challenges, we propose the Federated Local Data-Infused Graph Creation with Node-centric Model Refinement (Fed-LDR) algorithm. Fed-LDR leverages FL and Graph Convolutional Networks (GCN) to enhance spatio-temporal data analysis in urban environments. The algorithm comprises two key modules: (1) the Local Data-Infused Graph Creation (LDIGC) module, which dynamically reconfigures adjacency matrices to reflect evolving spatial relationships within urban environments, and (2) the Node-centric Model Refinement (NoMoR) module, which customizes model parameters for individual urban nodes to accommodate heterogeneity.
Evaluations on the PeMSD4 and PeMSD8 datasets demonstrate Fed-LDR's superior performance over six baseline methods. Fed-LDR achieved the lowest Mean Absolute Error (MAE) values of 20.15 and 17.30, and the lowest Root Mean Square Error (RMSE) values of 32.30 and 27.15, respectively, while maintaining a high correlation coefficient of 0.96 across both datasets. Notably, on the PeMSD4 dataset, Fed-LDR reduced MAE and RMSE by up to 81\% and 78\%, respectively, compared to the best-performing baseline FedMedian.
\end{abstract}

%

\section{Introduction}
Global urbanization is accelerating at an unprecedented rate, presenting complex challenges and opportunities for enhancing urban infrastructure and services. As cities expand and evolve, the environments within them become increasingly intricate, marked by dynamic interactions that traditional data types often fail to capture~\cite{thakuriah2017seeing}. In this context, spatio-temporal data—which integrates both spatial and temporal dimensions—has emerged as a critical tool for urban planners and policymakers. This data offers a nuanced understanding of urban phenomena, including traffic patterns, fluctuations in population density, and evolving environmental impacts over time~\cite{batty2013big,zheng2013u}. The effective utilization of spatio-temporal data is essential for informed decision-making aimed at promoting sustainability and resilience in urban development~\cite{Aburas2018Monitoring}.

Federated Learning (FL) has emerged as a promising paradigm, particularly suited for the Internet of Things (IoT) environments prevalent in urban settings. FL facilitates decentralized model training across diverse geographical locations, allowing the integration of various data sources while preserving data privacy~\cite{mcmahan2017communication,konevcny2016federated,gao2023pfdrl}. This decentralized approach not only enhances model flexibility but also enables personalized adjustments tailored to local conditions. Given the rapidly urbanizing environments, FL naturally aligns with the decentralized and heterogeneous nature of urban data, making it an ideal solution for developing adaptable and privacy-preserving models in real-world, complex urban landscapes.

While traditional computational models like regression analysis and agent-based models have been instrumental in understanding urban dynamics, integrating these models into the FL framework presents significant challenges, particularly in the analysis of spatio-temporal data within urban IoT environments. These traditional models are typically designed for analyzing static or relatively simple datasets, often focusing on specific variables or isolated snapshots in time. Although effective in certain contexts, they often fail to capture the intricate and evolving interdependencies of time and space that are inherent in urban systems~\cite{goodfellow2016deep}. Moreover, their static structure makes them ill-equipped to adapt to the continuous flux of urban life, thereby limiting their effectiveness in real-world applications~\cite{gao2021decentralized,yao2018modeling,liu2024fedbcgd,gao2022residential}.

Deep learning has revolutionized spatio-temporal data analysis by providing powerful tools for recognizing and learning complex patterns from large datasets~\cite{lecun2015deep}. However, significant challenges remain in integrating deep learning with FL for the analysis of spatio-temporal data within urban IoT environments. The primary challenge is developing models that not only achieve high accuracy but also adapt effectively to the unique characteristics of various urban nodes~\cite{zhang2016dnn, shi2016edge}. These nodes often exhibit substantial heterogeneity and evolving temporal dynamics, particularly in diverse urban settings. Consequently, the variability in local conditions can lead to suboptimal model performance~\cite{Fan2021Heterogeneous}.

\begin{figure}[t]
    \centering
    \begin{minipage}[t]{0.43\textwidth}
        \centering
        \includegraphics[width=\textwidth]{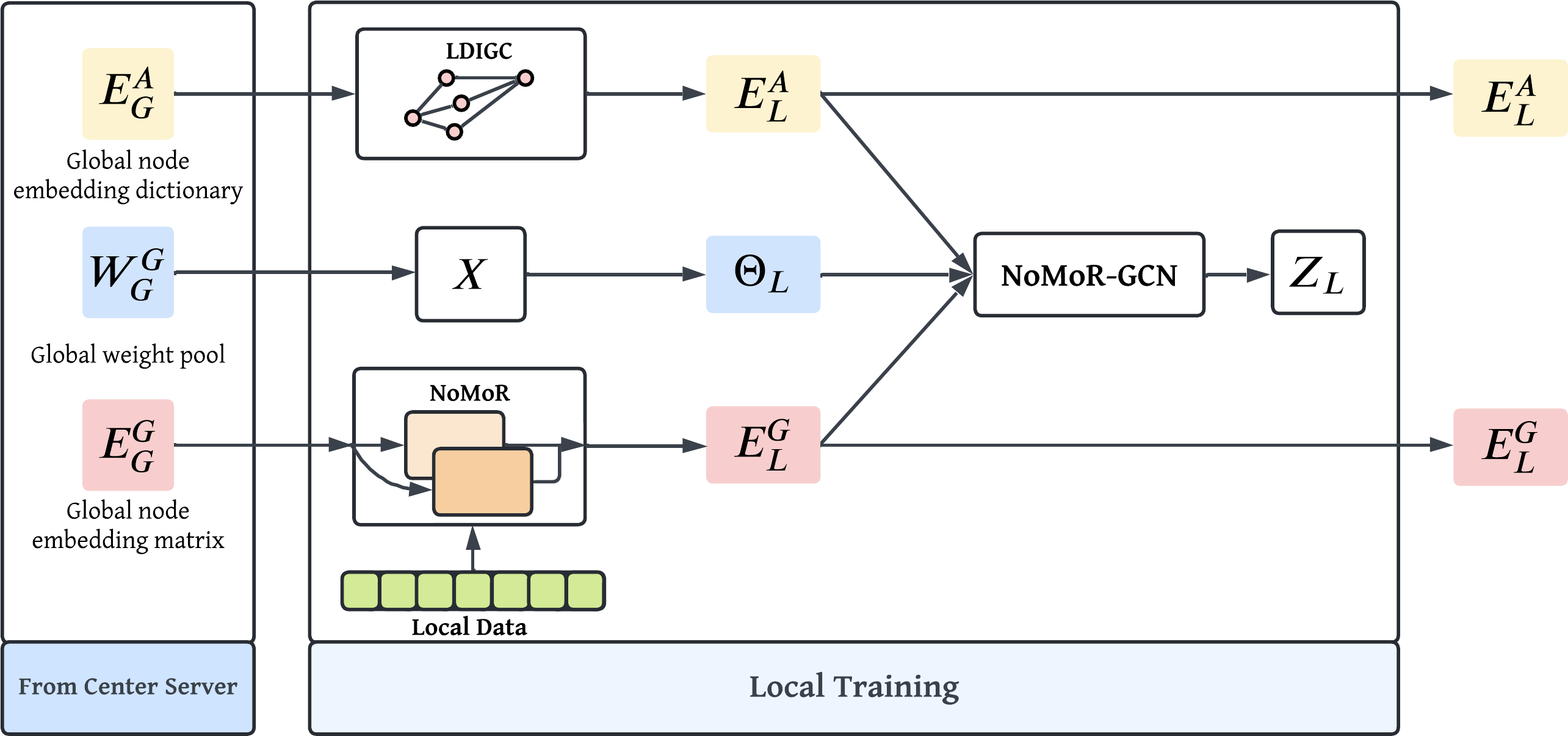}
        \caption{Client-side processing}
    \end{minipage}%
    \hfill
    \begin{minipage}[t]{0.43\textwidth}
        \centering
        \includegraphics[width=\textwidth]{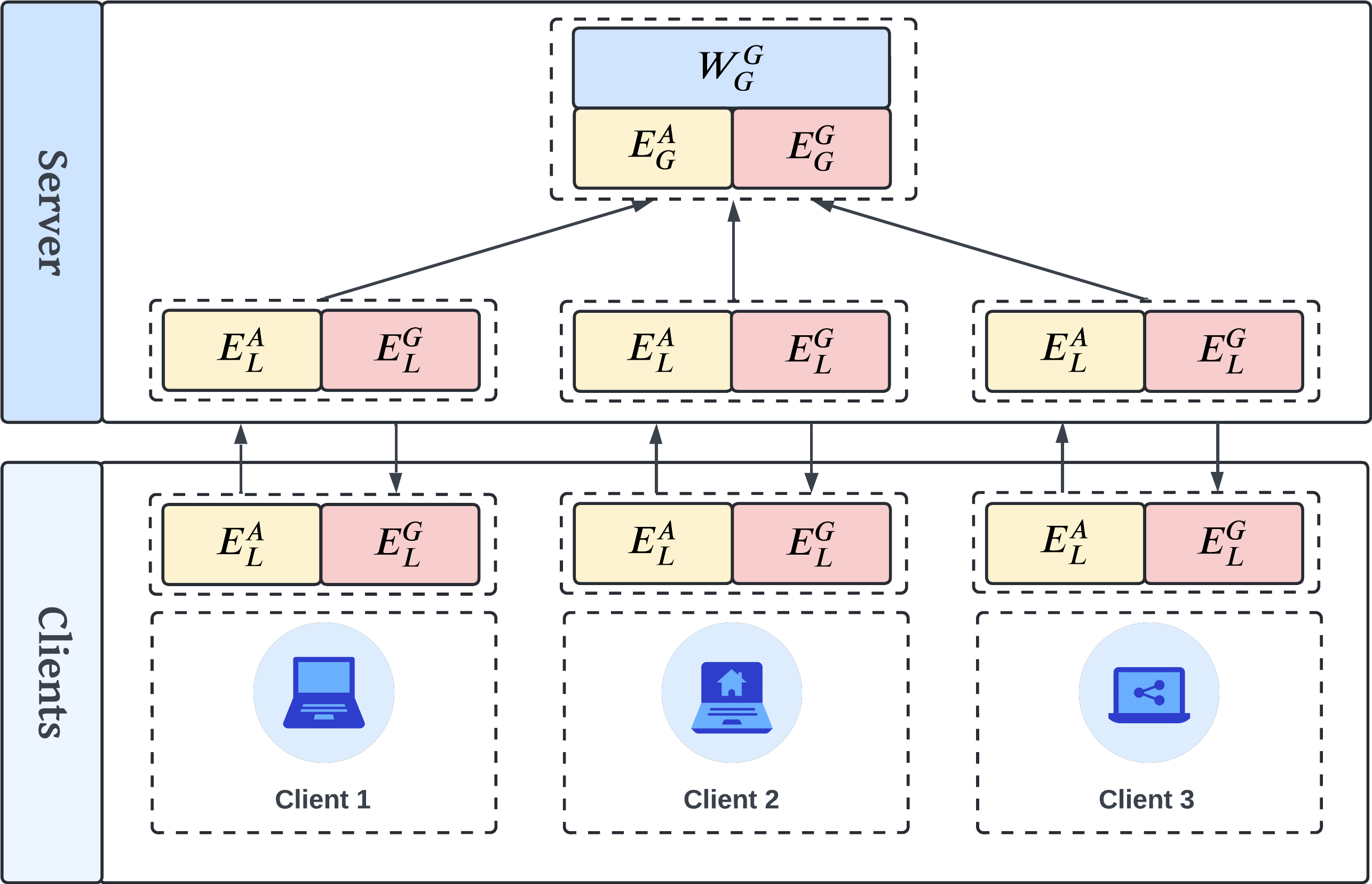}
        \caption{Central server aggregation}
    \end{minipage}
    \caption{Overview of our Proposed Fed-LDR.}
    \label{fig:overview}
\end{figure}

To address these challenges, we propose the Federated Local Data-Infused Graph Creation with Node-centric Model Refinement (Fed-LDR) algorithm. Fed-LDR combines the distributed learning strengths of FL with the spatial analysis capabilities of Graph Convolutional Networks (GCN), which excel at capturing spatial dependencies in spatio-temporal data~\cite{kipf2016semi}. As illustrated in Figure~\ref{fig:overview}, Fed-LDR comprises two key components: the Local Data-Infused Graph Creation (LDIGC) module and the Node-centric Model Refinement (NoMoR) module. LDIGC dynamically reconfigures adjacency matrices to reflect the continuous evolution of spatial relationships within urban environments, while NoMoR tailors model parameters for individual nodes, accommodating the heterogeneous nature of urban elements. This approach overcomes the limitations of static and centralized models, aligning with the inherently distributed and dynamic nature of urban data to enhance adaptability and protect privacy. By continuously adapting its learning framework based on localized, time-varying data inputs, Fed-LDR offers a more nuanced and context-aware analysis of urban dynamics, advancing the state-of-the-art in modeling complex, heterogeneous urban spatio-temporal phenomena.

Our comprehensive experiments on the PeMSD4 and PeMSD8 datasets demonstrate the superior performance of Fed-LDR compared to six baseline models, including traditional FL approaches and their variants enhanced with our proposed modules. Fed-LDR achieved the lowest Mean Absolute Error (MAE) and Root Mean Square Error (RMSE) on both datasets: MAE of 20.15 and RMSE of 32.30 for PeMSD4, and MAE of 17.30 and RMSE of 27.15 for PeMSD8. Notably, Fed-LDR maintained a high correlation coefficient (CORR) of 0.96 across both datasets, underscoring its robust predictive capabilities. These results represent significant improvements over traditional FL methods, with reductions in MAE and RMSE of up to 81\% and 78\%, respectively, compared to the best-performing baseline (FedMedian) on PeMSD4.

\section{Related Works}


\subsection{Spatio-Temporal Data Analysis with Graph Convolutional Networks}

In the realm of spatio-temporal data analysis using Graph Convolutional Networks (GCN), various works have laid the groundwork for understanding spatial dependencies within complex datasets. Studies like those by Li et al.~\cite{li2018dcrnn_traffic} and Yu et al.~\cite{yu2018spatio} initially paved the way for GCN integration into spatio-temporal forecasting, albeit with static adjacency matrices, which can limit the model's adaptability to evolving data patterns. Further, methods developed by Zhang et al.~\cite{zhang2020spatial} aimed to refine the adaptability of these models by incorporating attention mechanisms to dynamically weigh the importance of spatial relationships. Despite these advancements, the reliance on predefined spatial connections could overlook the intricate and evolving interactions characteristic of real-world scenarios.\looseness=-1

Recent research by Bai et al.~\cite{li2018adaptive} has attempted to make GCN more responsive to temporal dynamics, yet the persistent challenge of static spatial connections pointed out by Cui et al.~\cite{cui2019traffic} suggests a gap in fully capturing the discrete and dynamic roles of nodes within a network. This limitation often results in models that provide a generalized overview rather than a detailed depiction of individual node behaviors or temporal variations.
Wang et al.~\cite{wang2024tensor} introduced a tensor-based model that dynamically updates graph structures for enhanced spatio-temporal representation; Jain et al.~\cite{zainuddin2019improvement} developed a hybrid approach that improves temporal dynamics forecasting; and Guo et al.~\cite{guo2019attention} proposed a model employing attention mechanisms to adapt to temporal variations, collectively pushing the boundaries towards adaptive and nuanced analysis of spatio-temporal data. These approaches often fail to capture the discrete and evolving nature of each node's role within the network, leading to a generalized model that might not accurately reflect individual node behaviors or temporal variations.

\subsection{Federated Learning in Spatio-Temporal Analysis}

The field of federated learning in spatio-temporal analytics is relatively new, with only a handful of studies venturing into this domain. The work of STFL on vehicle trajectory prediction by leveraging federated learning emphasizes privacy while utilizing spatio-temporal traits to enhance prediction models, showcasing its effectiveness with the GAIA Open Dataset.~\cite{9998967} A parallel innovation for extensive wireless networks, also under the STFL banner, tackles intermittent learning and data delivery outages by introducing a compensatory mechanism, marking a stride in network resilience and learning performance.~\cite{6d1cb76a1415414081e75c3b3b2aa9c0}

Building upon these foundations, recent contributions have further explored the intersection of federated learning and spatio-temporal data analysis. For instance, the work of Wang et al.~\cite{wang2019adaptive}. introduces an adaptive federated learning framework tailored for urban traffic management systems, optimizing traffic flow predictions through decentralized data processing. . Similarly, Zhou et al.~\cite{huang2021starfl} have proposed a federated learning model that integrates real-time environmental monitoring data, facilitating localized pollution forecasts while maintaining user privacy.


Despite these advancements, these researches still grapple with the complexity of integrating highly variable spatial and temporal data, and the scalability of these models remains a challenge.
\section{Methodology}

\subsection{Problem Definition}

We address the problem of learning spatio-temporal dependencies in graph-structured data while preserving data privacy in a federated learning setting. The objective is to enhance predictive models for each node within a graph by leveraging both local data characteristics and global structural information without centralizing data storage. Consider a set of nodes \( \mathcal{N} = \{n_1, n_2, \ldots, n_i, \ldots, n_N\} \) where \( N \) is the total number of nodes, each associated with a temporal sequence of features. This sequence for a node \( n_i \) at time step \( t \) is represented by \( \mathbf{x}_{i,t} \in \mathbb{R}^F \), where \( F \) is the number of features. The problem is then to predict future feature vectors \( \mathbf{x}_{i,t+1}, \mathbf{x}_{i,t+2}, \ldots, \mathbf{x}_{i,t+\Delta} \) for a horizon of \( \Delta \) steps based on the historical observations \( \mathbf{x}_{i,t}, \mathbf{x}_{i,t-1}, \ldots, \mathbf{x}_{i,t-T+1} \), where \( T \) is the number of time steps in the past that are used for prediction.

The global structure of the system is represented by a graph \( G = (\mathcal{V}, \mathcal{E}, \mathbf{A}) \), where \( \mathcal{V} \) is the set of vertices corresponding to the nodes, \( \mathcal{E} \) is the set of edges representing the connections between nodes, and \( \mathbf{A} \in \mathbb{R}^{N \times N} \) is the adjacency matrix that encodes the proximity or similarity between nodes.

The challenge is to design a function \( \mathcal{F} \) parameterized by \( \Theta \), which accurately models the following relationship:

\[
\begin{aligned}
\{\mathbf{x}_{:,t+1}, \mathbf{x}_{:,t+2}, \ldots, \mathbf{x}_{:,t+\Delta}\} 
&=\\ \mathcal{F}_{\Theta}(\{\mathbf{x}_{:,t}, \mathbf{x}_{:,t-1}, \ldots, \mathbf{x}_{:,t-T+1}\}, G)
\end{aligned}
\]

The Federated Local Data-Infused Graph Creation with Node-centric Model Refinement (Fed-LDR) algorithm is proposed to learn the function \( \mathcal{F} \) in a federated setting, where each client improves a local graph-structured model based on its data while collaborating to refine a global model without direct data sharing. The algorithm's novelty lies in creating data-infused graph structures and performing node-centric model refinements to adapt the predictions to each node's specific characteristics, achieving high predictive accuracy while maintaining the privacy of the data across the network.

The aim is to iteratively update \( \Theta \) across a federation of clients while maintaining alignment with the global model, addressing challenges such as data heterogeneity, communication efficiency, and privacy concerns.

\subsection{Local Training with LDIGC and NoMoR}
\label{LDR}
The subsection presents Local Training by combining Local Data-Infused Graph Creation (LDIGC) and Node-centric Model Refinement (NoMoR). It showcases the benefit of refining models at a node level, which is crucial for detailed and strong system performance. We describe how each node uses local data to shape its computations. The nodes refine their models, focusing on unique data characteristics, which strengthens the overall system.

\subsubsection{Local Data-Infused Graph Creation (LDIGC)}

One of the challenges with current GCN-based traffic forecasting methodologies is the necessity for a predefined adjacency matrix \( A \) in the graph convolution procedure. Predominantly, these models employ either distance functions to determine the graph based on the geographical spacing between nodes or similarity functions to establish node proximity using node attributes (for instance, PoI data) or the traffic series itself. However, such methodologies, while intuitive, may not encapsulate the entirety of spatial dependencies and might not be directly aligned with predictive tasks, potentially leading to biases. Furthermore, the adaptability of these techniques to various domains without specific domain knowledge is limited, compromising the efficacy of conventional GCN-based models.

To address these challenges, we introduce the Local Data-Infused Graph Creation (LDIGC) module, engineered to automatically discern hidden interdependencies from the data. Initially, LDIGC sets up a learnable node embedding dictionary \( EA \in \mathbb{R}^{N \times d} \) for all nodes, where each row of \( EA \) symbolizes a node's embedding and \( d \) indicates the node embedding dimensionality. By leveraging the concept of node similarity, spatial dependencies between node pairs can be deduced via the matrix multiplication of \( EA \) and \( EA^T \):
\[ D^{-\frac{1}{2}}A D^{-\frac{1}{2}} = \text{softmax}(\text{ReLU}(EA \cdot EA^T)) \]
Here, the softmax function normalizes the adaptive matrix. Instead of generating \( A \) and then computing the Laplacian matrix, we produce \( D^{-\frac{1}{2}}A D^{-\frac{1}{2}} \) directly, minimizing redundant calculations in the iterative learning phase. As training progresses, \( EA \) undergoes updates to capture hidden traffic series dependencies, thus creating an adaptive matrix for graph convolutions. Compared to the self-adaptive adjacency matrix techniques, LDIGC is more straightforward and offers enhanced interpretability of the learned \( EA \). The overarching LDIGC-enhanced GCN can be described as:
\[ Z = (I_N + \text{softmax}(\text{ReLU}(EA \cdot EA^T)))X\Theta \]
In scenarios dealing with immense graphs (i.e., when \( N \) is substantial), LDIGC might impose significant computational demands. Strategies such as graph partitioning and sub-graph training can be utilized to mitigate this computational challenge.

\subsubsection{Node-centric Model Refinement (NoMoR)}
In recent developments in traffic forecasting, Graph Convolutional Networks (GCN) are primarily employed to discern the spatial correlations among traffic series, adhering to the computations delineated in the spectral domain. As suggested, the graph convolution operation is approximated using the 1st order Chebyshev polynomial expansion and can be broadened to high-dimensional GCN as:
\[ Z = \left( I_N + D^{-\frac{1}{2}}A D^{-\frac{1}{2}} \right) X \Theta + b \]
where \( A \) is the adjacency matrix of the graph, \( D \) represents the degree matrix, and \( X \) and \( Z \) signify the input and output of the GCN layer, respectively. The learnable weights and bias are symbolized by \( \Theta \) and \( b \). Focusing on a single node, say node \( i \), the GCN operation transitions the features of node \( X_i \) to \( Z_i \) utilizing the shared \( \Theta \) and \( b \) across nodes. Although this shared parameter methodology is conducive in recognizing dominant patterns across nodes in several problems, it appears suboptimal for intricate traffic forecasting challenges. Given the fluid nature of time-series data and distinct node-specific factors influencing traffic, we observe a multifaceted correlation landscape. For instance, neighboring nodes might exhibit disparate traffic patterns due to unique attributes, such as Points of Interest or weather conditions. Conversely, distant nodes could display contrasting traffic sequences. Hence, exclusively identifying mutual patterns across nodes isn't sufficient for nuanced traffic prediction, underscoring the need for individual parameter spaces tailored to each node.

However, establishing distinct parameters for every node culminates in an overwhelming parameter matrix \( \Theta \), which becomes challenging to optimize, particularly with a significant node count, and might induce overfitting. Addressing this, we introduce an enhancement to the traditional GCN through the Node-centric Model Refinement (NoMoR) module, drawing inspiration from matrix factorization techniques. Rather than directly deriving \( \Theta \), NoMoR focuses on learning two streamlined parameter matrices: a node-embedding matrix \( E_G \) and a weight pool \( W_G \). Subsequently, \( \Theta \) is generated as 
\[ \Theta = E_G \cdot W_G \]
Analyzing from the perspective of a specific node, such as node \( i \), this approach extracts parameters \( \Theta_i \) from a comprehensive shared weight pool \( W_G \) based on the node embedding \( E_{i}^G \). This procedure can be perceived as discerning node-specific patterns from a reservoir of potential patterns discerned from comprehensive traffic series. A parallel operation can be applied to \( b \). The final equation for the NoMoR-enhanced GCN can be articulated as:
\[ Z = \left( I_N + D^{-\frac{1}{2}}A D^{-\frac{1}{2}} \right) X E_G W_G + E_G b_G \]

\subsection{Fed-LDR }

 The Fed-LDR algorithm initializes at a central server by setting up a global node embedding dictionary, a node-embedding matrix, and a weight pool. These initial configurations are broadcast to all participating clients. On receiving these parameters, each client deploys the Local Data-Infused Graph Creation (LDIGC) module to deduce hidden data inter-dependencies, resulting in the refinement of the local node embedding dictionary. Subsequently, spatial dependencies are inferred using the updated local node embedding dictionary. Parallelly, the Node-centric Model Refinement (NoMoR) technique fine-tunes the model to individual nodes. This is achieved by utilizing the global node-embedding matrix to ascertain a local variant and subsequently deducing node-specific parameters. The local adaptation of the Graph Convolutional Network (GCN) model, termed NoMoR-GCN, is then executed using the local node-embedding matrix, the global weight pool, and local data. Post local adjustments, clients dispatch their updated node embedding dictionary and matrix to the central server. The central server then enhances the embedding dictionary and matrix from the clients with a regularization mechanism. This adjustment ensures local updates are closely aligned with the global model, effectively managing the challenges posed by data heterogeneity across clients. By integrating a similarity constraint into the optimization process of NoMoR-GCN, our method achieves a delicate balance between local data specificity and global model consistency, resulting in improved convergence and robustness in federated learning environments.



In the initial phase at the central server, a comprehensive setup is established. This involves the initialization of a global node embedding dictionary, denoted as $E_G^A \in \mathbb{R}^{N \times d}$, and a global node-embedding matrix, $E_G^G \in \mathbb{R}^{N \times d'}$. Additionally, a global weight pool $W_G^G \in \mathbb{R}^{d' \times C \times F}$ is created, where $d'$ signifies the embedding dimension pertinent to the NoMoR framework. These components are crucial as they are subsequently broadcasted to all participating client nodes, laying the groundwork for the algorithm's federated structure.

At the client level, each node commences with the received global parameters ($E_G^A$, $E_G^G$, and $W_G^G$) as the foundational base for local training. The clients utilize the LDIGC module, a pivotal element in the algorithm, to autonomously infer hidden inter-dependencies within their respective datasets. This step is crucial for updating the embedding dictionary to its local version, $E_L^A$, which in turn aids in inferring spatial dependencies. A significant aspect of this phase is the application of the NoMoR technique for node-specific model refinement. This involves leveraging the global node-embedding matrix $E_G^G$ to learn a local counterpart $E_L^G$, and subsequently deriving node-specific parameters through the formula $\Theta_L = E_L^G \cdot W_G^G$. The local execution of NoMoR-GCN is articulated as $Z_L = (I_N + D^{-\frac{1}{2}} A_L D^{-\frac{1}{2}}) X E_L^G W_G^G + E_L^G b^G$, representing a sophisticated computation within the client's domain.

Post the local training processes, each client forwards its updated node embedding dictionary ($E_L^A$) and node-embedding matrix ($E_L^G$) back to the central server. The central server plays a critical role in aggregating these individual updates to refine the global versions $E_G^A$ and $E_G^G$. This aggregation process may adopt a weighted average approach, factoring in the size or quality of each client's data, to ensure a more effective and representative global model. 

The algorithm is characterized by its iterative nature, where the steps of local training and central aggregation are repeated over multiple communication rounds. This iterative process is maintained until the algorithm converges or reaches a pre-determined number of rounds, ensuring thoroughness and accuracy in the model's development.

\begin{enumerate}
    \item \textbf{Global Node Embedding Dictionary \(E^A_G\) (from the LDIGC module)}:
    \begin{itemize}
        \item This refers to the embeddings used in the Local Data-Infused Graph Creation (LDIGC) process. In the original LDIGC description, the embeddings are utilized to automatically infer the hidden inter-dependencies in the graph.
        \item These embeddings help in constructing an adaptive adjacency matrix to represent the graph structure, which is based on the data itself.
        \item It's designed to address that traditional adjacency matrices (either based on distances or similarity functions) may not capture all relevant spatial dependencies.
    \end{itemize}

    \item \textbf{Global Node-Embedding Matrix \(E^G_G\) (from the NoMoR module)}:
    \begin{itemize}
        \item This is associated with the Node-centric Model Refinement (NoMoR) concept, which introduces a mechanism for nodes to have their distinct model parameters.
        \item The idea is to extract node-specific patterns from a set of candidate patterns discovered from all traffic series.
        \item The matrix \(E^G_G\) represents node-specific embeddings, which, when combined with a weight pool \(W^G_G\), generate node-specific model parameters.
    \end{itemize}
\end{enumerate}

In simpler terms, while both embeddings are concerned with nodes, the purpose they serve and the modules they originate from are different. \(E^A_G\) is used to generate an adaptive graph structure (i.e., adjacency matrix) based on data, while \(E^G_G\) is used to adaptively generate node-specific model parameters for better forecasting. Algorithm~\ref{Algo} shows the detailed structure of our proposed Fed-LDR. 

\begin{algorithm}
\caption{Fed-LDR: Federated Local Data-infused Graph Creation with Node-centric Model Refinement}
\label{Algo}
\begin{algorithmic}[1]

\Procedure{Initialize Central Server}{}
    \State $E_G^A \gets \text{InitializeMatrix}(N, d)$
    \State $E_G^G \gets \text{InitializeMatrix}(N, d')$
    \State $W_G^G \gets \text{InitializeMatrix}(d', C, F)$
    \State \text{Broadcast} $E_G^A, E_G^G, W_G^G$ \text{to all clients}
\EndProcedure

\Procedure{LocalTraining}{Client}
    \State \text{Receive} $E_G^A, E_G^G, W_G^G$ \text{from server}
    \State $E_L^A \gets \text{Employ LDIGC using local data and } E_G^A$
    \State \text{Infer spatial dependencies using} $E_L^A$
    \State $E_L^G \gets \text{Learn node-embedding using } E_G^G$
    \State $\Theta_L \gets E_L^G \times W_G^G$
    \State $Z_L \gets (I_N + D^{-1/2} \times A_L \times D^{-1/2}) \times X \times E_L^G \times W_G^G + E_L^G \times b^G$
    \State \Return $E_L^A, E_L^G$
\EndProcedure

\Procedure{Aggregation}{AllClientsData}
    \For{\text{each client's data in} AllClientsData}
        \State \text{Aggregate} $E_L^A$ \text{and} $E_L^G$ \text{to update} $E_G^A$ \text{and} $E_G^G$
    \EndFor
\EndProcedure

\Procedure{Fed-LDR}{}
    \For{rounds = 1 \text{to} maxRounds \textbf{or} until convergence}
        \State AllClientsData $\gets$ \text{EmptyList}
        \For{each client}
            \State ClientData $\gets$ \Call{LocalTraining}{Client}
            \State \text{Add} ClientData \text{to} AllClientsData
        \EndFor
        \Call{Aggregation}{AllClientsData}
    \EndFor
\EndProcedure

\end{algorithmic}
\end{algorithm}

\section{Experiments}
\subsection{Datasets}

For the evaluation of the Fed-LDR algorithm, we conducted our experiments on 2 datasets that are characteristic of spatio-temporal graph-structured data:

\textbf{PeMSD4}: This dataset comprises traffic flow data from the California Department of Transportation's Performance Measurement System (PeMS), focusing on the San Francisco Bay Area. It includes information from 307 loop detectors, spanning the period of January 1, 2018, to February 28, 2018. The data is utilized to model and predict traffic patterns in a federated learning setting.
  
\textbf{PeMSD8}: Similar to PeMSD4, the PeMSD8 dataset is also sourced from PeMS but concentrates on the San Bernardino area. It captures the traffic flow data collected from 170 loop detectors over the period from July 1, 2016, to August 31, 2016. This dataset helps in understanding the dynamics of a different urban traffic environment under federated learning paradigms.
  
  

\begin{table*}
\centering
\resizebox{0.9\textwidth}{!}{%
\begin{tabular}{lcccccccc}
\toprule
\textbf{Method} & \multicolumn{4}{c}{\textbf{PEMSD4}} & \multicolumn{4}{c}{\textbf{PEMSD8}} \\
\cmidrule(r){2-5} \cmidrule(r){6-9}
 & \textbf{MAE} & \textbf{RMSE} & \textbf{RCC} & \textbf{Corr} & \textbf{MAE} & \textbf{RMSE} & \textbf{RCC} & \textbf{Corr} \\
\midrule
FedOpt      & 124.80 & 167.26 & 1.17 & 0.40 & 128.42 & 164.33 & 1.60 & 0.14  \\
FedAvg      & 118.33 & 156.45 & 1.12 & 0.41 & 112.09 & 144.39 & 1.25 & 0.53 \\
FedMedian   & 108.69 & 144.18 & 1.04 & 0.52 & 132.96 & 169.34 & 1.48 & 0.23 \\
\midrule
LDIGC+NoMoR+FedOpt          & 20.25  & 32.20  & 0.21 & 0.96 & 18.00 & 27.91 & 0.19 & 0.96 \\
LDIGC+NoMoR+FedAvg          & 20.36  & 32.53  & 0.21 & 0.96 & 17.47 & 27.25 & 0.19 & 0.96 \\
LDIGC+NoMoR+FedMedian       & 21.53  & 33.45  & 0.22 & 0.96 & 17.82 & 27.66 & 0.19 & 0.96 \\
LDIGC+NoMoR (Locally)        & 24.22  & 42.32  & 0.29 & 0.96 & 22.84 & 38.52 & 0.29 & 0.95 \\
\midrule
Fed-LDR (Ours)         & 20.15  & 32.30  & 0.21 & 0.96 & 17.30 & 27.15 & 0.19 & 0.96 \\ 
\bottomrule
\end{tabular}%
}
\caption{Performance Metrics Comparison of Federated Learning Models Across the Different Datasets} 
\label{performance_table}
\end{table*}

\subsection{Data Preprocessing and Experiment Setups}

Each dataset underwent rigorous preprocessing to ensure consistency and quality, including cleaning, normalization, and aggregation of traffic flow readings into meaningful intervals. 


For each dataset, a graph was constructed to reflect the spatial dependencies inherent in the data. The  PeMSD4 and PeMSD8 naturally lent itself to a grid-like graph structure. Each node was then associated with its corresponding time series data, and edge weights were computed using domain-specific similarity measures. To simulate the federated learning environment, each dataset was split into multiple client datasets. These splits were constructed to reflect the distribution of data across different geographical locations for the traffic datasets, and across different camera views for the re-identification datasets.

For the training and validation process, we adopted a standard 70\%-15\%-15\% train-validation-test split. Temporal continuity was preserved in the split to maintain the integrity of the spatio-temporal patterns crucial for our algorithm's efficacy. We conduct all of our experiments on the A100 GPUs in Google Colab Pro Plus.

\subsection{Baselines}

\textbf{Base Model - GCN}: This model builds upon the Diffusion Convolutional Recurrent Neural Network (DCRNN)~\cite{li2018diffusion} and incorporates a single-layer Graph Convolutional Network (GCN) within the DCRNN gates to generate feature embeddings. The encoder is composed of multiple layers of GCN-DCRNN units, with two layers typically yielding the best performance. This architecture effectively captures complex spatiotemporal patterns and produces predictions through a single convolutional block decoder.

To evaluate our proposed method, we compare it against the following six baseline methods, all utilizing the aforementioned GCN as the base model:

\begin{itemize}
\item \textbf{FedAvg}~\cite{mcmahan2017communication}: A federated learning algorithm that averages model updates from multiple clients to improve a global model.
\item \textbf{FedMedian}~\cite{yin2018byzantine}: A variant of federated learning that uses the median of client updates instead of the average for increased robustness.
\item \textbf{FedOpt}~\cite{asad2020fedopt}: An advanced federated learning approach that incorporates adaptive optimization techniques to handle client heterogeneity.
\item \textbf{LDIGC+NoMoR+FedAvg}: Combines local data infusion and node-specific optimization with FedAvg for federated learning.
\item \textbf{LDIGC+NoMoR+FedMedian}: Integrates local data infusion and node-specific optimization with FedMedian for robust federated learning.
\item \textbf{LDIGC+NoMoR+FedOpt}: Merges local data infusion and node-specific optimization with FedOpt's adaptive capabilities for enhanced federated learning.
\end{itemize}

\section{Experimental Result}

For evaluating the performance of our innovative Fed-LDR algorithm, we have chosen three widely recognized indicators: Mean Absolute Error (MAE), Root Mean Square Error (RMSE), and Mean Absolute Percentage Error (MAPE). These indicators together provide a holistic measure of our model's predictive precision. In Table \ref{performance_table}, we present a consolidated view of the performance metrics for our algorithm in comparison to 4 established methods.

Our performance analysis utilizes a diverse set of metrics tailored to spatio-temporal data analysis: Mean Absolute Error (MAE), Root Mean Square Error (RMSE), Mean Absolute Percentage Error (MAPE), along with Correlation (CORR). These metrics, evaluated across multiple prediction horizons for the PeMSD4 and PeMSD8 datasets, offer insights into both the temporal accuracy and spatial consistency of our model. 



\subsection{Performance Analysis}

\begin{figure*}[ht]
    \centering
        \includegraphics[width=\textwidth]{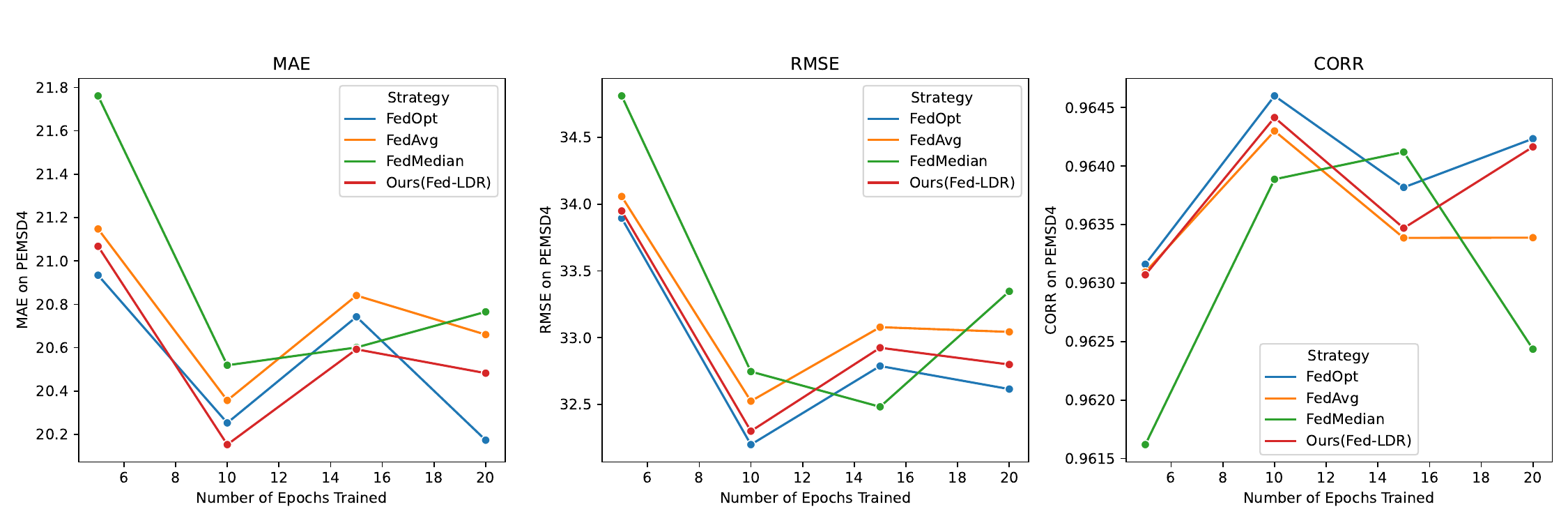}
    \caption{Influence of Local Epoch Number on Model Performance on PeMSD04.}
    \label{epoch}
\end{figure*}

In Table \ref{performance_table}, we observe the comparative performance of the Fed-LDR model against established federated learning strategies over datasets PeMSD4 and PeMSD8. The Fed-LDR outperforms others with consistently lower MAE and RMSE values, achieving 20.15 and 32.30 on PeMSD4 and 17.30 and 27.15 on PeMSD8, respectively. It maintains a higher CORR across horizons, indicating not only better accuracy but also a more reliable prediction capability. These findings highlight the superiority of Fed-LDR's node-centric refinement in handling various prediction horizons without the performance deterioration seen in other models.

\begin{figure*}[ht]
\centering
\includegraphics[width=\textwidth]{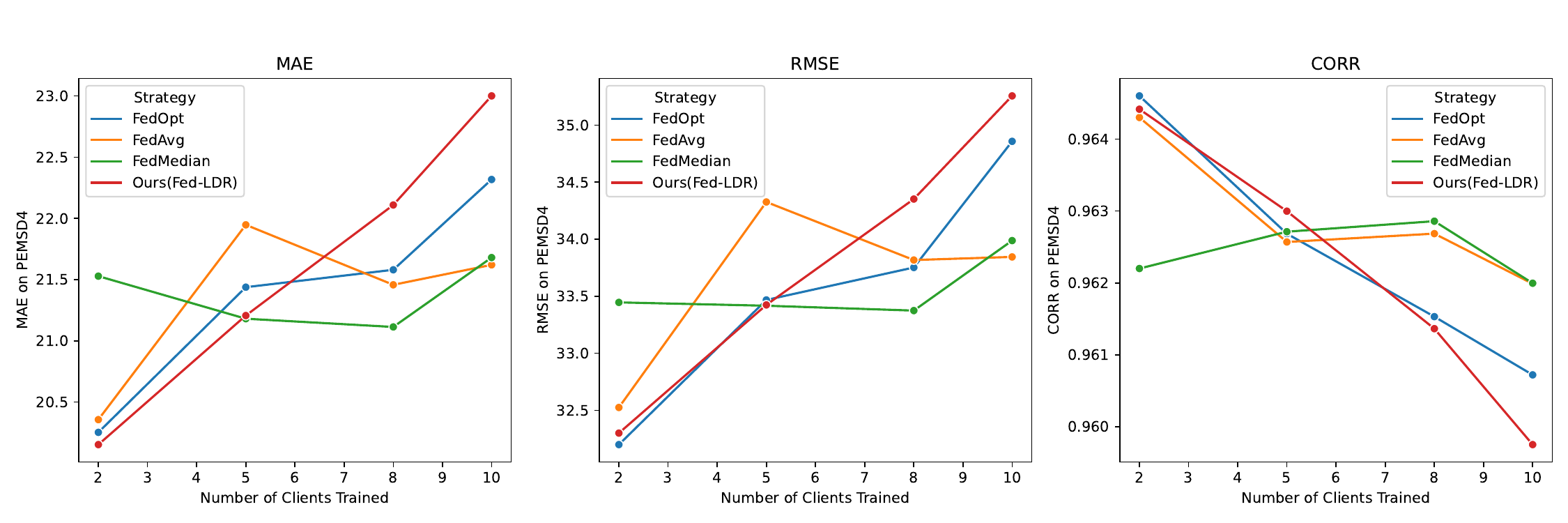}
\caption{Influence of the Number of Clients on Model Performance on PeMSD04.}
\label{fig:client_PEMDs4}
\end{figure*}

\subsection{Different Local Epochs}

In our study, the number of local epochs—periods during which the model learns from its local data before updating the global model—plays a significant role in performance. Figure \ref{epoch} compares the influence of local epoch number on model performance metrics (MAE, RMSE, CORR) for PeMSD04 dataset.

Figure~\ref{epoch} indicates a clear trend: as the number of epochs increases, the error rates (MAE, RMSE) tend to first decrease, showing an improvement in the model's predictive capabilities. However, this trend reverses after a certain number of epochs, suggesting that too many local epochs may lead to overfitting to the local data and a decrease in overall model performance.

Regarding the correlation (CORR), the optimal number of epochs is the point where the model's predictions are most aligned with the actual data. Here, the figure suggests that fewer epochs might not capture enough of the data's nuances, while too many epochs may start to capture noise, leading to less reliable predictions.

Different federated strategies also show varying optimal epochs, which highlights the importance of tuning the local epoch number in accordance with the federated learning strategy employed. Balancing the number of epochs is crucial: not enough and the model doesn't learn adequately, too many and the model's generalizability is compromised.

\subsection{Different Number of Clients}

Figure \ref{fig:client_PEMDs4} shows how the number of clients impacts the performance of federated learning models. As the number of clients increases from 2 to 6, the evaluation metrics opt for better model accuracy due to diverse data. However, beyond 6 clients, MAE and RMSE increase, indicating a decline in performance, possibly due to data inconsistency. The Correlation (CORR) peaks around 4 to 6 clients, after which it slightly decreases. These patterns suggest an optimal client range exists that maximizes model performance without introducing excessive data variability.

The exclusion of the PeMSD8 dataset from the "Different Number of Clients" experiment was a strategic decision, primarily influenced by the aim to ensure a focused and manageable analysis within the constraints of computational resources and the complexity inherent in this kind of experiments. The federated learning framework, especially in scenarios involving a significant increase in the number of clients, demands substantial computational resources. Given the larger scale and increased granularity of the Pemsd08 dataset, extending this experiment to include it would have exponentially amplified the computational burden. This consideration was pivotal, as our objective was to delineate clear, actionable insights within feasible computational and time constraints. The primary aim of this section was to elucidate the impact of client diversity on model accuracy and performance metrics. The Pemsd04 dataset, with its well-defined and consistent traffic flow patterns, provided a robust platform for this investigation. By concentrating on this dataset, we were able to more effectively isolate and examine the influence of client numbers, ensuring that our conclusions were both robust and directly attributable to the variables in question.



\section{Conclusion}
Our proposed Fed-LDR algorithm integrated federated learning with graph convolutional network, has proven to be a pivotal stride towards understanding and leveraging the intricate dynamics of spatio-temporal data. Through innovative approaches to local data integration and model refinement, this study demonstrates enhanced model accuracy and efficiency across diverse network environments. The findings underscore the importance of optimizing client participation and highlight the potential of Fed-LDR in addressing data heterogeneity and scalability challenges. Future work will explore broader applications and further refine the algorithm to maximize federated learning's benefits across various domains. This research lays a solid foundation for the next generation of federated learning models, promising improved performance and adaptability.




\bibliographystyle{IEEEtran}
\bibliography{main}

\end{document}